\documentclass[conference, table]{IEEEtran}
\IEEEoverridecommandlockouts

\usepackage[pdfstartview=XYZ,
bookmarks=true,
colorlinks=true,
linkcolor=blue,
urlcolor=blue,
citecolor=blue,
pdftex,
bookmarks=true,
linktocpage=true, 
hyperindex=true
]{hyperref}

\usepackage{orcidlink}

\usepackage{balance}
\usepackage{epstopdf}
\usepackage{booktabs}
\usepackage[utf8]{inputenc}
\usepackage[T1]{fontenc}
\usepackage[linesnumbered,ruled,vlined, noend]{algorithm2e}
\let\oldnl\nl
\newcommand{\nonl}{\renewcommand{\nl}{\let\nl\oldnl}}

\SetCommentSty{mycommfont}
\usepackage{multirow}
\usepackage[abbreviations]{foreign}
\usepackage[inline]{enumitem}
\usepackage{cite}
\usepackage{amsmath,amssymb,amsfonts,mathtools}
\usepackage{comment}
\newcolumntype{P}[1]{>{\centering\arraybackslash}p{#1}}
\newcolumntype{M}[1]{>{\centering\arraybackslash}m{#1}}
\usepackage{listings}
\usepackage{caption}
\usepackage{svg}
\usepackage{tabularx, makecell, booktabs}
\newcommand\mydots{\makebox[0.8em][c]{.\hfil.\hfil.}}
\AtBeginDocument{

  \newcommand{\sys}{\textsc{Pelta}\xspace}
  \definecolor{darkgreen}{rgb}{0.3,0.5,0.3}
  
  \definecolor{darkred}{rgb}{0.5,0.3,0.3}

}

\usepackage{framed}
\usepackage[binary-units,per-mode=symbol]{siunitx}

\usepackage{hyperref}
\hypersetup{
 hidelinks = true,
    colorlinks = false,
    linkbordercolor = {white}
}

\definecolor{darkgreen}{rgb}{0.3,0.5,0.3}
\definecolor{darkblue}{rgb}{0.3,0.3,0.5}
\definecolor{darkred}{rgb}{0.5,0.3,0.3}

\newboolean{showcomments}
\setboolean{showcomments}{false}
\ifthenelse{\boolean{showcomments}}{ 
  \newcommand{\mynote}[3]{
    \fbox{\bfseries\sffamily\scriptsize#1}
    {\small$\blacktriangleright$
     \textsf{\emph{\color{#3}{#2}}}$\blacktriangleleft$
     }
   }
}{\newcommand{\mynote}[3]{}}
\newcommand{\sq}[1]{\mynote{simon}{#1}{orange}}
\newcommand{\vs}[1]{\mynote{valerio}{#1}{red}}

\def\BibTeX{{\rm B\kern-.05em{\sc i\kern-.025em b}\kern-.08em
    T\kern-.1667em\lower.7ex\hbox{E}\kern-.125emX}}
\begin{document}

\title{Mitigating Adversarial Attacks in Federated Learning with Trusted Execution Environments}



\author{\IEEEauthorblockN{Simon Queyrut~\orcidlink{0000-0002-1354-9604},
Valerio Schiavoni~\orcidlink{0000-0003-1493-6603}, Pascal Felber~\orcidlink{0000-0003-1574-6721}}
\IEEEauthorblockA{Institute of Computer Science (IIUN)\\
University of Neuchâtel,
Neuchâtel, Switzerland\\ \texttt{first.last@unine.ch}}
}


\maketitle

\begin{abstract}
The main premise of federated learning (FL) is that machine learning model updates are computed locally to preserve user data privacy.
This approach avoids by design user data to ever leave the perimeter of their device.
Once the updates aggregated, the model is broadcast to all nodes in the federation. 
However, without proper defenses, compromised nodes can probe the model inside their local memory in search for adversarial examples, which can lead to dangerous real-world scenarios. 
For instance, in image-based applications, adversarial examples consist of images slightly perturbed to the human eye getting misclassified by the local model.
These adversarial images are then later presented to a victim node's counterpart model to replay the attack. 
Typical examples harness dissemination strategies such as altered traffic signs (patch attacks) no longer recognized by autonomous vehicles or seemingly unaltered samples that poison the local dataset of the FL scheme to undermine its robustness.
\sys is a novel shielding mechanism leveraging Trusted Execution Environments (TEEs) that reduce the ability of attackers to craft adversarial samples.
\sys masks inside the TEE the first part of the back-propagation chain rule, typically exploited by attackers to craft the malicious samples.
We evaluate \sys on state-of-the-art accurate models using three well-established datasets: CIFAR-10, CIFAR-100 and ImageNet.
We show the effectiveness of \sys in mitigating six white-box state-of-the-art adversarial attacks, such as Projected Gradient Descent, Momentum Iterative Method, Auto Projected Gradient Descent, the Carlini \& Wagner attack. In particular, \sys constitutes the first attempt at defending an ensemble model against the Self-Attention Gradient attack to the best of our knowledge.
Our code is available to the research community at \url{https://github.com/queyrusi/Pelta}.
\end{abstract}

\section{Introduction}
The proliferation of edge devices and small-scale local servers available off-the-shelf nowadays generated an astonishing trove of data, to be used in several areas, including smart homes, e-health, \etc.
For several of these scenarios, the data being generated is highly sensitive.
While the deployment of data-driven machine learning (ML) algorithms to train models over such data is becoming prevalent, one must take special care to prevent privacy leaks.
In fact, it has been shown how, without proper mitigation mechanisms, sensitive data (\ie the one used by such ML during training) can be reconstructed.
To overcome this problem, an increasingly popular approach is federated learning (FL)~\cite{45648,kairouz2021advances}.
FL is a decentralized machine learning paradigm, where clients share with a trusted server only their local individual updates, rather than the data used to train it, hence protecting by design the privacy of user data.
The trusted FL server is known by all nodes.
His role is to build a global model by aggregating the updates sent by the nodes.
Once aggregated, the server broadcasts back the updated model to all clients.
The nodes will update their models locally and use the following updates with a fresh batch of local data (\ie for inference purposes).
This approach prevents user-data from leaving the user devices, as only the local model updates are sent outside the device.

A popular model trained in FL is the transformer~\cite{NIPS2017_3f5ee243}, a widespread multi-purpose deep learning (DL) architecture. Transformers create a rich and high-performing continuous representation of the whole input (\ie text or images~\cite{Raganato2018AnAO}), effectively used across a diverse range of tasks, yielding state-of-the-art results in several areas, \eg language modeling~\cite{Shoeybi2019, brown2020language}, translation~\cite{DBLP:journals/corr/abs-2104-06022}, audio classification~\cite{NEURIPS2021_76ba9f56}, computer vision tasks such as image classification with vision transformer models (ViT)~\cite{50650}, object detection \cite{wei2022FD}, semantic segmentation \cite{beit}, \etc.
They harness the \emph{self-attention mechanism}~\cite{NIPS2017_3f5ee243},
which weights the relative positions of tokens inside a single input sequence to compute a representation of that given sequence. 

In the FL training context, both the fine tuning of transformer-based models pre-trained on large-scale data and the training of their lightweight counterparts such as MobileViT~\cite{mehta2022mobilevit}
have sparked interest in industry and academia~\cite{ro2022scaling}.
Because this family of models demands considerable efforts in design and computing resources, protecting its integrity during a collaborative training requires special attention. 

Consider now the following conjectured scenario.
A server broadcasts its expensive model to collaborating clients to have it finetuned over their local private data.
One of the clients taps into its RAM, copies the model and computes malicious patches, specifically designed to trick the model. Without ever altering the model, he can now run a patch attack~\cite{Brown2017AdversarialP}: he puts adversarial stickers on objects (roadsigns for instance) that are subject to regular inferences by the FL model: the objects are then misclassified by unaware agents running the collaboratively learned model and an accident may ensue.

Alternatively, the malicious agent initiates a poisoning attack that can break a model’s robustness by sending the central server updates that stem from inference on samples engineered with a trojan trigger to create an unsuspected backdoor~\cite{Bagdasaryan2020Jun} that can be activated at an inconvenient time by unaware users of the FL model.
Similarly, malicious clients can have the model purposefully and repeatedly misclassify their newfound adversarial examples to severely undermine the quality of the aggregated updates~\cite{Bhagoji2019May}. In all these scenarios, the malicious client accessing its own physical copy of the model allowed him to generate poisonous data that effectively compromised the FL model's reliability.
Safeguarding against the creation of these adversaries is paramount in a framework where scaling amplifies the effectiveness of attacks to alarming levels.

\begin{figure}[!t]
    	\begin{center}
		\includegraphics[scale=0.6,trim={0 4 0 0}]{{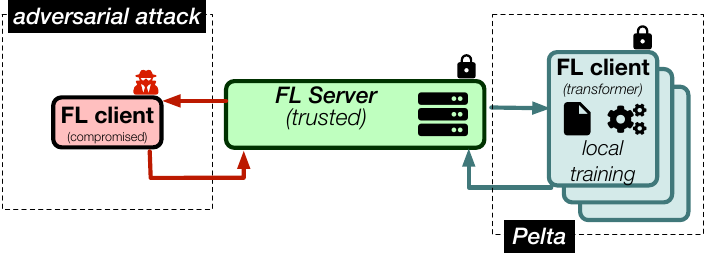}}
    	\end{center}
	\caption{\label{fig:fl}Federated learning under trusted and compromised nodes. \sys shields against evasion attacks.\vs{The adversarial image can potentially be broadcast}}
\end{figure}

In this work, we focus on a client crafting adversarial examples. Fig.~\ref{fig:fl} depicts our considered FL scenario with a compromised node which tries to craft adversarial  samples (\ie launching an evasion attack).
Because it is the hardest to defend against, we consider the \emph{white-box} setting, \ie the model's characteristics are completely known, and an attacker can leverage gradient computation inside the model to craft adversarial examples.
For instance, in the case of vision classification, an attacker leverages the model's gradients to craft images designed to fool a classifier while appearing seemingly unaltered to humans, launching a so-called gradient-based \emph{adversarial} (or \emph{evasion}) \emph{attack}~\cite{DBLP:journals/corr/GoodfellowSS14, 10.1007/978-3-642-40994-3_25, DBLP:journals/corr/SzegedyZSBEGF13}.\footnote{While we use Vision Transformers for illustration and description purposes, the mitigation mechanisms presented and implemented later are directly applicable to other classes of models including DNNs or Transformers, such as those for NLP, audio processing, \etc.}
By design of the FL paradigm, a large number of compromised client can probe their own device memory to launch an adversarial attack against the FL model. 
In a white-box scenario, these attacks exploit the model in clear at inference time, by perturbing the input and having it misclassified. Such perturbations are typically an additive mask crafted by following the collected gradients of the loss \emph{w.r.t} the input pixels and applying it to the input signal. Such gradients are either directly obtained by tapping into the device's memory when they are effectively computed for local usage or they can be calculated knowing model weights and activations when the device is instructed not to produce them (this is typically the case when running inferences after deployment).

In this paper, we propose \sys: it is a defense that mitigates gradient-based adversarial attacks launched by a client node by leveraging hardware obfuscation at inference time to hide a few in-memory values, \ie those close to the input and produced by the model during each pass. 
This leaves the attackers unable to complete the chain-rule of the back-propagation algorithm used by gradient-based attacks and compute the perturbation to update his adversarial sample.
To this end, we rely on hardware-enabled trusted execution environments (TEE), by means of enclaves, secure areas of a processor offering privacy and integrity guarantees.
Notable examples include Intel SGX~\cite{costan2016intel} or Arm TrustZone~\cite{amacher2019performance}. 
TEEs can lead to a restriced white-box scenario, \ie, a stricter setting in which the attacker cannot access absolutely everything from the model he seeks to defeat,
hence impairing his white-box attack protocols designed for the looser hypothesis.
In our context, we deal specifically with TrustZone, given its vast adoption, performance~\cite{amacher2019performance} and support for attestation~\cite{menetrey2022watz}.
However, TrustZone enclaves have limited memory (up to 30~MB in some scenarios), making it challenging to completely shield the state-of-the-art Transformer architectures often larger than 500~MB. This constraint therefore calls for such a hardware defense to be a light, partial obfuscation of the model.

Our main contributions are as follows:
\begin{enumerate}[label=(\roman*)]
    \item We show that it is possible to mitigate evasion attacks during inference in a FL context through hardware shielding.
	\item We describe the operating principles of \sys, our lightweight gradient masking defense scheme.
    \item We apply \sys to shield layers of individual and ensemble state-of-the-art models against several white-box attack including the Self-Attention Gradient Attack to show the scheme effectively provides high protection in a nigh white-box setting.
    \item To the best of our knowledge, \sys is the first applied defense against the Self-Attention Gradient Attack.
	\vs{\item  update list}
\end{enumerate}

The rest of the paper is as follows. 
\S\ref{section:related} surveys related work.
\S\ref{section:threat} presents our threat model. 
\S\ref{section:design} describes the principles of the \sys shielding defense. 
We evaluate it on an ensemble model against a gradient-based attack and discuss the results in \S\ref{section:eval}. 
\S\ref{section:discuss} discusses general system implications of \sys . 
Finally, we conclude and hint at future work in \S\ref{section:conclu}.
\section{Related Work}
\label{section:related}
A significant body of work exists towards defending against adversarial attacks in a white-box context~\cite{CarliniSurvey}. 
Because of its few-constraints hypothesis, particular endeavour is directed towards refining adversarial training (AT) methods~\cite{DBLP:journals/pr/QianHWZ22, DBLP:journals/corr/abs-2102-01356}.
However, recent studies show a trade-off between a model's generalization capabilities (\ie, its standard test accuracy) and its robust accuracy~\cite{raghunathan*2019adversarial, Nakkiran2019AdversarialRM, tsipras2018robustness, Chen2020MoreDC}. 
AT can also expose the model to new threats~\cite{https://doi.org/10.48550/arxiv.2202.10546} and, perhaps even more remarkably, increase robust error at times~\cite{DBLP:journals/corr/abs-2203-02006}. 
It is possible to use AT at training time as a defense in the federated learning context~\cite{10.48550/ARXIV.2203.04696} but it creates its own sensitive exposure to a potentially malicious server. 
The latter can restore a feature vector that is tightly approximated when the probability assigned to the ground-truth class is low (which is the case for an adversarial example)~\cite{DBLP:conf/iclr/ZhangCSBDH19}. 
Thus, the server may reconstruct adversarial samples of its client nodes, successfully allowing for privacy breach through an \emph{inversion attack}, \ie, reconstructing elements of private data in other nodes' devices.
Surprisingly, \cite{10.48550/ARXIV.2203.04696} also uses randomisation at inference time~\cite{10.5555/3454287.3454433} to defend against iterative gradient-based adversarial attacks, even though much earlier work expresses worrying reserves about such practice~\cite{DBLP:conf/icml/AthalyeC018}. 

In FL, where privacy of the users is paramount, defending against inversion is not on par with the security of the model in itself and attempts at bridging the two are ongoing~\cite{liu2022threats}. 
These attempts focus on defending the model at training time or against \emph{poisoning}, \ie, altering the model's parameters to have it underperfom in its primary task or overperform in a secondary task unbeknownst to the server or the nodes. 
On the other hand, defenses against inversion attacks have seen a surge since the introduction of FL. 
DarkneTZ~\cite{10.1145/3386901.3388946}, PPFL~\cite{10.1145/3458864.3466628} and GradSec~\cite{Messaoud2022Nov} mitigate these attacks with the use of a TEE.
By protecting sensitive parameters, activations and gradients inside the enclave memory, the risk of gradient leakage is alleviated and the white-box setting is effectively limited, thus weakening the threat.
However, \sys is conceptually different from these methods, as it does not consider the gradient of the loss with respect to the \emph{parameters}, but with respect to the \emph{input image} instead. 
The former can be revealing private training data samples in the course of an inversion attack, while the latter only ever directs the perturbation applied to an input when conducting an adversarial attack. 
Other enclave-based defenses do not focus on adversarial attacks, are based on SGX (meaning looser enclave constraints) and do not deal with ML computational graphs in general~\cite{10.1145/3488659.3493779, Hanzlik2021Jun}, yet present defense architectures of layers that fit our case~\cite{DBLP:journals/corr/abs-1807-00969}.
Where those were tested only for CNN-based architectures or simple DNNs, \sys can protect a larger, more accurate Transformer-based architecture. 
The robustness of the two types against adversarial attacks was compared in prior studies~\cite{DBLP:journals/corr/abs-2110-02797, 9710333, DBLP:journals/corr/abs-2106-03734}.

In~\cite{Zhang2021Nov}, authors mitigate inference time evasion attacks by pairing the distributed model with a node-specific \emph{attractor}, which detects adversarial perturbations, before distributing it. 
However, \cite{Zhang2021Nov} assumes the nodes only have black-box access to their local model.
Given the current limitations on the size of encrypted memory in particular for TrustZone enclaves~\cite{amacher2019performance}, it is currently unfeasible to completely shield models such as VGG-16 or larger. 
This is why \sys aims at exploring a more reasonable use case of the hardware obfuscation features of TrustZone by shielding only a subset of the total layers, as we detail more later.
It could thus be used as an overlay to the aforementioned study. More generally, our proposed defense scheme does not interfere with existing software solutions for train time or inference time defenses such as randomization, quantization or encoding techniques~\cite{REN2020346}.
As a result, \sys should not be regarded as a competitor algorithm when it comes to obfuscating sensitive gradients but rather as a supplementary hardware-reliant aid to existing protocols.

Overall, gradient obfuscations as a defense mechanism against evasion attacks have been studied in the past~\cite{DBLP:conf/icml/AthalyeC018, tramer2018ensemble}. 
Authors discuss from a theoretical perspective the fragility of relying on masking techniques.
Instead, we show that it fares well protecting a state-of-the-art architecture against inference time gradient-based evasion attacks even by using off-the-shelf hardware with limited resources. 
We further elaborate on the strong ties to the Backward Pass Differentiable Approximation (BPDA) attack~\cite{DBLP:conf/icml/AthalyeC018} in \S\ref{section:design}. While we show that \sys unambiguously weakens a malicious agent in the white-box setting by restricting their lever of performance, its design provides no defense capabilities against black-box attacks~\cite{Wang2022Aug} since they operate in a setting that already assumes complete obfuscation of the model’s quantities for crafting the adversarial samples. 

\section{Threat Model}
\label{section:threat}
We assume an honest-but-curious client attacker, which does not tamper with the FL process and message flow.
The attacker's device is assumed to be computing the gradients of the model at inference time, which is typically the case at each inference during the training rounds of a FL scheme.
The case where no gradients are produced/stored by the device is a subcase of this setting. 
The normal message exchanges defined by the protocol are not altered. 
The attacker has access to the model to run inferences, but a subset of the layers are \emph{shielded}, under a restricted white-box setting ensured by an impregnable TEE enclave (side channel attacks are out of scope for the rest of this paper). 
In practice, secure communication channels are established so that only privileged users can allow data recovery from within the TEE.
A layer $l$ is shielded (as opposed to its normal \textit{clear} state) if some variables and operations that directly lead to computing gradient values from this layer are obfuscated, \ie hidden from an attacker.
In a typical deep neural network (DNN) $f$ such that 
$$f=\operatorname{softmax} \circ f^n \circ f^{n-1} \circ \cdots \circ f^1$$
with layer $f^i=\sigma^i(W^i\cdot x+b^i)$ ($\sigma^i$ are activation functions), this implies, from shallow (close to the input) to deep (close to loss computation):
 the input of the layer $a^{l-1}$;
 its weight $W^{l}$ and bias $b^{l}$;
 its output $z^{l}$ and parametric transform $a^{l}$.
In general terms, a layer encompasses at least one or two transformations.
The attacker probes its own local copy of the model to search for adversarial examples, ultimately to present those to victim nodes and replicate the misclassification by their own copy of the model. 
\section{Design}
\label{section:design}
We explain first the design of \sys shielding of a model, with details on the general mechanisms and the shielding approach in \S\ref{ssec:peltashield}.
Then, we confront \sys to the case of the BPDA~\cite{DBLP:conf/icml/AthalyeC018} attack in \S\ref{subsection:bpda}.
\subsection{Back-Propagation Principles and Limits}
Recall the back-propagation mechanism; consider a computational graph node $x$, its immediate (\ie, generation $1$) children nodes $\{u^1_j\}$, and the gradients of the loss function with respect to the children nodes $\{{d \mathcal{L}}/{d u^1_j}\}$.
The back-propagation algorithm uses the chain rule to calculate the gradient of the loss function $\mathcal{L}$ with respect to ${x}$ as in
\begin{equation}
\label{eqn:backprop}
    \frac{d \mathcal{L}}{d {x}}=\sum_j\left(\frac{\partial f^1_j}{\partial {x}}\right)^T \frac{d \mathcal{L}}{d u^1_j}
\end{equation}
where $f^1_j$ denotes the function computing node $u^1_j$ from its parents $\alpha^1$, \ie $u^1_j=f^1_j\left(\alpha^1\right)$.
In the case of Transformer encoders, $f^i_j$ can be a convolution, a feedforward layer, an attention layer, a layer-normalization step, \etc.
Existing shielded white-box scenarios~\cite{Messaoud2022Nov} protect components of the gradient of the loss \emph{w.r.t.} the parameters $\nabla _{\theta}\mathcal{L}$, to prevent leakage otherwise enabling an attacker to conduct inversion attacks.
Instead, we seek to mask gradients that serve the calculation of the gradient of the loss \emph{w.r.t.} the \emph{input} $\nabla _{x}\mathcal{L}$ to prevent leakage enabling gradient-based adversarial attacks, as those exploit the backward pass quantities (\ie, the gradients) of the input to maximise the model's loss.

Because the model shared between nodes in the FL group still needs to back-propagate correct gradients at each node of the computational graph, masking only an intermediate layer is useless, since it does not prevent the attacker from accessing the correct in-memory gradients on the clear left-hand (shallow) side of the shielded layer. 
We thus always perform the gradient obfuscation of the first trainable parameters of the model, \ie, its shallowest successive layers: it is the lightest way to alter $\nabla _{x}\mathcal{L}$ through physical masking.
Specifically, using Eq.~\ref{eqn:backprop}, where $x$ denotes the input image (\ie, the adversarial instance $x_{adv}$), an attacker could perform any gradient-based evasion attack (a special case where a layer is not differentiable is discussed in \S\ref{subsection:bpda}). 
\sys renders the chain rule incomplete, by forcibly and partially masking the left-hand side term inside the sum, $\{\partial f_{j}^{1}/\partial {x}\}$, hence relying on the lightest possible obfuscation to prevent collecting these gradients and launch the attack. 
\subsection{\sys shielding}\label{ssec:peltashield}
We consider the computational graph of an ML model:
$$\mathcal{G}=\left\langle n, l, E, u^1 \ldots u^n, f^{l+1} \ldots f^n\right\rangle$$ where $n$ and $l$ respectively represent the number of nodes, \ie, transformations, and the number of leaf nodes (inputs and parameters) \emph{s.t.} $1 \leq l<n$. 
$E$ denotes the set of edges in the graph. 
For each $(j,i)$ of $E$, we have that $j<i$ with $j\in\{1 \mydots(n-1)\}$ and $i\in\{(l+1)\mydots n\}$. The $u^i$ are the variables associated with every numbered vertex $i$ inside $\mathcal{G}$. They can be scalars or vector values of any dimension. The $f^i$ are the differentiable functions associated with each non-leaf vertex.
We also assume a TEE enclave $\mathcal{E}$ that can physically and unequivocally hide quantities stored inside (\eg side-channel to TEEs are out of scope).

To mitigate the adversarial attack when gradients are stored in memory, the \sys shielding scheme presented in Alg.~\ref{algo:proc} stores in the TEE enclave $\mathcal{E}$ all the children jacobian matrices of the first layer, $\{\partial f_{j}^{1}/\partial {x}\}$.
\begin{algorithm}[!t]
  \SetAlgoLined\DontPrintSemicolon\SetNoFillComment
  \SetKwFunction{algo}{algo}\SetKwFunction{proc}{proc}
  \SetKwProg{myalg}{Algorithm}{}{}
  \KwData{$\mathcal{G}=\langle n, l, E, u^1 \ldots u^n, f^{l+1} \ldots f^n\rangle$; enclave $\mathcal{E}$}
  \SetKwProg{myproc}{Algorithm}{}{}
  \nonl\myproc{\DataSty{\sys}$(\mathcal{G})$}{
  $S \gets$ \DataSty{Select}$(u^{l+1} \ldots u^n)$ \label{line:1}\;
  \lFor{$u$ in $S$}
        {\DataSty{Shield}$(u, \mathcal{E})$}
   \KwRet\;}
   \nonl\myalg{\DataSty{Shield$(u^i, \mathcal{E})$}}{
   $\mathcal{E} \gets \mathcal{E}+\{u^i\}$ \label{line:6}\;
   $\alpha^i \gets \langle u^j \mid(j, i) \in E \rangle$\tcp*[f]{get parent vertices}\;
   \For{$u^j$ in $\alpha^i$} {\If{$u_j$ is input \label{line:9}}
   {$J^{j \rightarrow i} \gets {\partial f^i\left(\alpha^i\right)}/{\partial u^j}$ \tcp*[h]{local jacobian}\label{line:10}\;
    $\mathcal{E} \gets \mathcal{E}+\{J^{j \rightarrow i}\}$ \label{line:11}\;
    }
    \DataSty{Shield}$(u^j,\mathcal{E})$ \tcp*[r]{move on to parent}}
  \KwRet \label{line:13} \;}{}
  \caption{\sys shielding}
  \label{algo:proc}
\end{algorithm} 
 Note that, given the summation in Eq.~\ref{eqn:backprop}, obfuscating only some of the children jacobians $\partial f_1^1 / \partial x$, $\partial f_2^1 / \partial x \mydots$ would already constitute an alteration of $\nabla_x \mathcal{L}$. 
We however chose that \sys masks all the partial factors of this summation to not take any chances.
This obfuscation implies that intermediate gradients should be masked as well. Indeed, because they lead to $\partial f_{j}^{1}/\partial {x}$ through the chain rule, the intermediate gradients (or \textit{local jacobian}) $J^{0 \rightarrow 1}=\partial f^1\left(\alpha^1\right) / \partial u^0$ between the input $x=u^0$ of the ML model and its first transformation must be masked~(Alg.~\ref{algo:proc}-line \ref{line:11}). 
Because one layer may carry several 
transforms on its local input (\eg linear then $\operatorname{ReLU}$), local jacobians should be masked for at least as many subsequent generations of children nodes as the numbers of layers to shield.
The number of generations is directly determined at the arbitrary selection step~(Alg.~\ref{algo:proc}-line \ref{line:1}) where the defender choses how far the model should be shielded, \ie, the deepest masked nodes. 
In practice, selecting the first couple of nodes likely induces enough alteration to mitigate the attack and prevent immediate reconstruction of the hidden parameters through either direct calculus (input-output comparison) or inference through repeated queries~\cite{Oh2019Sep}. 
From this deep frontier, the defender may recursively mask the parent jacobians ~(Alg.~\ref{algo:proc}-line \ref{line:11},~\ref{line:13}). Note that this step is skipped in practice when the device doesn't store any gradients.

In adversarial attacks, the malicious user keeps the model intact and treats only the input $x_{adv}$ as a trainable parameter to maximize the prediction error of the sample.
We note that the local jacobians~(Alg.~\ref{algo:proc}-line \ref{line:10}) between a child node and their non-input parents need not be hidden because the parents are not trainable (they should be input of the model to meet the condition at Alg.~\ref{algo:proc}-line \ref{line:9}).
Such quantities are, in effect, simply not present in the back-propagation graph and could not constitute a leak of the sensitive gradient.
Similarly, notice how \textsc{Select} supposes the deepest masked nodes be chosen \emph{s.t.} they come \textit{after} every input leaf nodes (\ie, from subsequent generations). This condition: $u^i\in S\Rightarrow i>l$, insures no information leaks towards trainable input leaf nodes.
After all sensitive partials are masked by Alg.~\ref{algo:proc}, such a set of obfuscated gradients is a subset of the total forward partials that would lead to a complete chain rule and is noted $\left\{\partial f/\partial {x}\right\}^L$ assuming $L\leq n$ is the deepest vertex number that denies the attacker information to complete the chain rule.
Since all the forward partials down to $L$ are masked, then $\partial f^L /\partial {x}$ is protected and the resulting under-factored gradient vector (\ie the \textit{adjoint} of $f^{L+1}$) is a vector in the shape of the shallowest clear layer $f^{L+1}$ and noted $\delta_{L+1}=d \mathcal{L} /d {u^{L+1}}$. 
A white-box setting assumes an attacker has knowledge of both the subset $\left\{\partial f/\partial {x}\right\}^L$ of all forward partials and $\delta_{L+1}$ to perform a regular gradient-based update on his $x_{adv}$.
However in \sys, the attacker is only left with the adjoint $\delta_{L+1}$.

Finally, the forward pass quantities $u^i$ that may lead to the unambiguous recovery of the hidden set $\left\{\partial f/\partial {x}\right\}^L$ are masked (Alg.~\ref{algo:proc}-\ref{line:6}).
This insures that arguments $\alpha^i$ of the transformations $f^i$ cannot be further exploited by the attacker. 
This could happen when one node is a linear transform of the other as in $u^{i+1}=f^{i+1}((W,u^{i}))=W\times u^{i}$.
In this case, the local jacobian $J^{i \rightarrow i+1}$ is known to be exactly equal to $W$.
Notice that weights and biases of a DNN would be effectively masked, as they are regarded as leaf vertices of the model's computational graph for the $f^i((u^{i}, u^{i-1}, u^{i-2}))=u^{i-1}\cdot u^{i} + u^{i-2}=Wx + b$ operation. Similarly, the outputs of the transformations, $u^i=f^i(\alpha^i)$, are masked by (Alg.~\ref{algo:proc}-line \ref{line:6}). 
As an example, for a DNN, the exact quantities enumerated in \S\ref{section:threat}
are stored in the enclave $\mathcal{E}$. 

Overall, \sys should be construed as simply the general principle of unequivocally hiding enough parameters and gradients that are directly next to the input so that they cannot be maliciously exploited.
\subsection{Relation with BPDA}
\label{subsection:bpda}
\begin{figure}[!t]
    	\begin{center}
		\includegraphics[scale=0.6,trim={0 0 0 0}]{{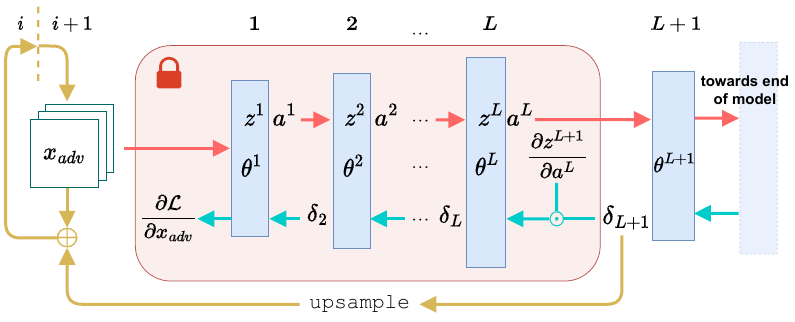}}
    	\end{center}
	\caption{Overview of the \sys defense scheme over the first layers of a machine learning model against an iterative gradient-based adversarial attack.
	Because the attacker cannot access operations in the shallow layers, he resorts to upsample his under-factored gradient $\delta_{L+1}$ to compute the adversarial update.
	The figure depicts activations as transforms for a DNN.  \label{fig:shieldnn}}
\end{figure}

When training a DL model, a straight-through estimator allows a simple approximation of the gradient of a non-differentiable function (\eg a threshold)~\cite{DBLP:journals/corr/BengioLC13}.
In a setting discussed in~\cite{DBLP:conf/icml/AthalyeC018}, this idea is generalized to a so called Backward Pass Differentiable Approximation (BPDA) to illustrate how preventing a layer from being differentiable in hopes of defeating a gradient-based attack actually constitutes a fragile security measure.
In BPDA, the non-differentiable layer $f^l$ of a neural network $f^{1 \ldots L}(\cdot)$ would be approximated by a neural network $g$ \emph{s.t.} $g(x) \approx f^l(x)$ and be back-propagated through $g$ instead of the non-differentiable transform.
This method is what a malicious node adopts against \sys by upsampling the adjoint of the last clear layer $\delta_{L+1}$ to bypass the shielded layers. An illustrative case for a DNN is shown in Fig.\ref{fig:shieldnn}. However, the attacker operates with two limiting factors: \textit{(i)} in a real-world scenario, the attacker possesses limited time and number of passes before the broadcast model becomes obsolete; \textit{(ii)} we hypothesize the attacker does not have priors on the parameters of the first layers of the model, therefore the adversary is effectively left without options for computing the gradient-based update other than training a BPDA of the layer.
Although this attack makes a fundamental pitfall of gradient masking techniques, it is worth noting this step becomes increasingly difficult for the attacker as larger parts of the model are hidden from him since it would suppose he has training resources equivalent to that of the FL system. As a side note, we mention that there exist recent defenses against BPDA~\cite{9420266}.

In \S\ref{section:eval}, we investigate to what extent an adversarial attack can be mitigated by \sys and whether the upsampling of the under-factored gradient (which is merely a linear transformation of the correct gradient in the case of the first transformation in many vision models) as a substitute for the masked backward process constitutes a possible last resort for the malicious node.
\section{Evaluation}
\label{section:eval}
\vs{\textit{What is the effect of forcibly removing leftmost factors from the chain-rule of an attacker conducting an adversarial attack?}}
\emph{Does \sys mitigate white-box attacks?}
To answer this question, we conduct attacks on several models with shielded layers.
\subsection{ Evaluation Metric and Ensemble Defense Setup}
In a common classification task, the \textit{clean accuracy} of a model refers to its standard test accuracy to differentiate it from the context of an adversarial attack where the goal of the defender is to score high on \textit{astuteness} (\ie \textit{robust} accuracy) against a set of correctly classified samples to which adversarial perturbations were added. This means that a perfectly \emph{astute} model would almost always correctly classify a perturbed sample if he classified it correctly when it was clean.
Against adversarial attacks, we chose as defending models several state-of-the-art high clean accuracy models trained on three heavily benchmarked image datasets: CIFAR-10, CIFAR-100~\cite{Krizhevsky09} and the ImageNet-21K dataset~\cite{Russakovsky2015Dec}. 
\subsubsection{Individual defenders} 
\label{subsubsection:indiv}
Because of the widespread use of the attention mechanism in a large variety of machine learning tasks, we included three size variants of the Vision Transformer in our experiments. 
Specifically: ViT-L/16, ViT-B/16 and ViT-B/32~\cite{50650}.
We also included two conventional CNNs, namely ResNet-56, ResNet-164~\cite{He2016Sep} and two Big Transfer models: BiT-M-R101x3 and BiT-M-R152x4~\cite{10.1007/978-3-030-58558-7_29} which stem from the CNN-based ResNet-v2 architecture~\cite{He2016Sep}.
\subsubsection{Ensemble defender}
\label{subsubsection:ensemble}
Additionally to these individual defending models we also study the astuteness of an \emph{ensemble} of a ViT and a BiT. An ensemble of models consists of two or more models that determine the correct output through a decision policy.
We chose an ensemble because, generally, when dealing with the image classification task, adversarial examples do not \textit{transfer} well between attention based and CNN based models~\cite{9710333}.
This means that an example crafted to fool one type of model in particular will rarely defeat the other, thus highly benefiting the astuteness of the ensemble.
This allows for mitigating attacks that target either one of the two specifically, by exploiting a combination of the model outputs to maximize chances of correct prediction. 
In this paper, we use random selection~\cite{Srisakaokul2018Aug} as a decision policy, where, for each sample, one of the two models is selected at random to evaluate the input at test time. 
\begin{center}
\begin{table}[!t]
	\centering
	\normalsize
	\setlength{\tabcolsep}{7.63pt}
	\begin{center}
		\rowcolors{1}{gray!0}{gray!10}
		\begin{tabularx}{\columnwidth}{lcc}
			\rowcolor{gray!50}
			\textbf{Model} & \textbf{Shielded portion} & \textbf{TEE mem. used}\\
			\rowcolor{gray!1}
			ViT-L/16  &   1.34\% & 15.16~MB \\
                ViT-B/16  &   3.61\% & 11.97~MB \\
			BiT-M-R101x3   & $4.50\mathrm{e}{-3}$\% & 65.20~KB \\
                BiT-M-R152x4   & $9.23\mathrm{e}{-3}$\% & 322.14~KB \\
			
		\end{tabularx}
    \rule{250pt}{0.7pt}
	\end{center}
	\caption{\label{tab:size}Estimated enclave memory cost and model portion shielded in each setting.
	The ensemble value sums both models in the worst case where enclaves are not flushed between evaluation of either of the two models.}
	\label{table:size}
\end{table}
\end{center}
\textbf{\sys Shielding defense of the white-box.}
To simulate the shielding inside the TEE enclave, we deprive the attacker of the aforementioned quantities (\S\ref{ssec:peltashield}) during the individual attacks (\ref{subsubsection:indiv}) and during the attack against the ensemble  (\ref{subsubsection:ensemble}). To the best of our knowledge, this is the first ever attempt at mitigating the recent Self-Attention Gradient Attack (see \ref{subsection:saga}).  Against the ensemble, the attacker is deprived of the sensitive quantities of the two models separately, then jointly. 
For the ViT models, all transforms up to the position embedding~\cite{50650} are included. 
This means that the following steps occur inside the enclave: separation of the input into patches ${x}_p^n$, projection onto embedding space with embedding matrix $E$, concatenation with learnable token $x_{\text {class}}$ and summation with position embedding matrix $E_{\text{pos}}$:
$$
z_0=\left[x_{\text {class }} ; {x}_p^1 {E} ; {x}_p^2 {E} ; \cdots ; {x}_p^N {E}\right]+{E}_{\text {pos}}
$$

For the Big Transfer (BiT) models, the scheme includes the first weight-standardized convolution~\cite{10.1007/978-3-030-58558-7_29} and its following padding operation. For the ResNets, the first convolution, batch normalization and ReLU activation are masked.
Notice that, for all three model types, we obfuscate either two learnable transformations or a non-invertible parametric transformation like weight-standardization, ReLU or MaxPool~\cite{Zeiler2014}, so the attacker cannot retrieve the obfuscated quantities without uncertainty.
Table~\ref{table:size} reports the estimated overheads of the shield for each setting: the theoretical memory footprints of each secured intermediate activation, weight and gradient as single-precision floating-point numbers were summed and are shown for the ImageNet dataset variants of the models in the worst case where intermediate activations and gradients inside the shield are not flushed after the back-propagation algorithm uses them to complete the pass.
Assuming the most resource-intensive case where gradients are produced, the shielding of the ensemble requires less than 16~MB of TEE memory at the very worst, consistent with what typical TrustZone-enabled devices allow~\cite{amacher2019performance}. Notice that, because it only ever obfuscates the shallowest parts of a model, \sys is barely ever affected by the scale of larger and more complex variants, which makes it suitable for a wide variety of state-of-the-art models.

\subsection{Attacker Setup} 
\label{subsection:saga}

\begin{figure}[!t]
    	\begin{center}
		\includegraphics[scale=1,trim={0 0 0 0}]{{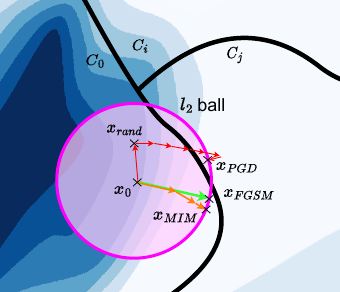}}
    	\end{center}
	\caption{
            Schematic diagram of three gradient-based maximum allowable adversarial methods. Within a norm constraint, the attacker computes an additive perturbation of input $x_0$ to cross a decision boundary of a victim model. Only PGD (red) was able to craft an adversarial example $x_{PGD}$ here.
            \label{fig:gradientMethods}}
\end{figure}

\sq{Add a short description of the 5 attacks. Equations...? !! THIS IS JUST A BLUNT COPY PASTE FOR NOW}
Against the individual models (\ref{subsubsection:indiv}), we launch four iterative maximum allowable attacks and one regularization-based attack. Against the ensemble model (\ref{subsubsection:ensemble}) we launch one iterative maximum allowable attack.
We shortly introduce all six in their non-targeted version here, \ie, the specific class which the altered sampled is misclassified into has no importance.
For the sake of conciseness, we omitted some descriptive equations that can be found in the original papers of the considered attacks.

Iterative maximum allowable attacks (Fig.~\ref{fig:gradientMethods}) start at an initial point $x^{(0)}$ which can be chosen as the origin sample $x_0$ or sometimes as a randomly sampled $x_{rand}$.
The attack then iterates adversarial candidates $x_{adv}^{(i)}$ by following an overall ascendant path of the loss function of the model within a norm constraint. This implies adversarial samples are required to stay inside an $l_2$ or $l_{\infty}$ ball centered on $x_{0}$, \ie, $||x_{adv}^{(i)}-x_0||_{2}$ or $||x_{adv}^{(i)}-x_0||_{\infty} \le \epsilon, \forall i\geq0$.
In the case of $l_{\infty}$, this means that the features (in this paper, the individual pixels of the image) cannot vary more than $\pm\epsilon$ in magnitude.
Specifically, we use the following attacks in this paper:

\textbf{Fast Gradient Sign Method} - The Fast Gradient Sign Method (FGSM)~\cite{DBLP:journals/corr/GoodfellowSS14} is a one-step gradient-based approach that finds an adversarial example $x_{adv}$ by adding a single $\epsilon$-perturbation that maximizes the loss function $\mathcal{L}$ to the original sample of label $y$ such that $x_{adv}=x_0+\epsilon \cdot \operatorname{sign}\left(\nabla_x\mathcal{L}(x_0, y)\right)$.

\textbf{Projected Gradient Descent} - The Projected Gradient Descent (PGD)~\cite{Madry2018} is the natural multi-step variant of the FGSM algorithm that insures the resulting adversarial example $x_{adv}$ stays within bound of a maximum allowable $\epsilon$-perturbation.
This is done through a $P$ operator that projects out of bound values back into the $\epsilon$-ball as illustrated by the last PGD step of Fig.~\ref{fig:gradientMethods}.
The $i^{\text{th}}$ step is computed as $x^{(i)}=P\left(x^{(i-1)}+\epsilon_{\text{step}} \cdot \operatorname{sign}\left(\nabla_x \mathcal{L}(x_0, y)\right)\right)$, $\epsilon_{\text{step}}$ being the step size.

\textbf{Momentum Iterative Method} - Inspired by the popular momentum method for accelerating gradient descent algorithms, the Momentum Iterative Method (MIM)~\cite{Dong2018Jun} applies a velocity vector in the direction of the gradient of the loss function across iterations and at any step $i>0$ takes into account previous gradients ($g^{(i)}_\mu$ relies on $g^{(i-1)}_{\mu}$) with a decay factor $\mu$.
The additive step is reminiscent of FGSM, as the $i^{\text{th}}$ step is computed by $x^{(i)}=x^{(i-1)}+\epsilon_{\text{step}} \cdot \operatorname{sign}(g^{(i)}_\mu)$.

\textbf{Auto Projected Gradient Descent} - The Auto Projected Gradient Descent (APGD)~\cite{Croce2020Jul} proposes adaptive changes of the step size in PGD.
This automated scheme allows for an exploration phase and an exploitation phase where an objective function (commonly the cross-entropy loss) is maximized.
This attack includes other mechanisms like the ability to restart at a best point along the search. In the benchmark of the individual models, APGD is the most recent attack and it would also be considered the most sophisticated.

We also use a so-called \emph{regularization-based} attack on our individual defending models~\ref{subsubsection:indiv}:

\textbf{Carlini and Wagner Attack} - The Carlini and Wagner Attack (C\&W)~\cite{7958570} iteratively minimizes (typically through the gradient descent algorithm) an objective sum of two competing terms.
Through various variable substitutions, one term indirectly measures the norm (typically, $l_2$) of the added perturbation.
On the other hand, a so-called \emph{regularized} term evaluates the wrongness of the classification for an adversarial candidate $x_{adv}$.

Lastly, we present an iterative gradient-based method against our ensemble defense:

\textbf{Self-Attention Gradient Attack} - To circumvent the ensemble defense~\ref{subsubsection:ensemble}, the attacker uses a  gradient-sign attack: the Self-Attention Gradient Attack (SAGA)~\cite{9710333}.
It iteratively crafts the adversarial example by following the sign of the gradient of the losses as follows:
\begin{equation}
\label{eq:fgsm}
    x^{(i+1)}=x^{(i)}+\epsilon_{\text{step}} * \operatorname{sign}\left(G_{b l e n d}\left(x^{(i)}\right)\right)
\end{equation}
where we initialize $x^{(0)}=x_0$ the initial image and $\epsilon_{\text{step}}$ is a given set attack step size (chosen experimentally).
Additionally, $G_{b l e n d}$ is defined as a weighted sum, each of the two terms working towards computing an adversarial example against either one of the two models:
\begin{equation}
    G_{b l e n d}\left(x^{(i)}\right)= \alpha_{k} \frac{\partial \mathcal{L}_{k}}{\partial x^{(i)}}+ \alpha_{v} \phi_{v} \odot \frac{\partial  \mathcal{L}_{v}}{\partial x^{(i)}}
\end{equation}
$\partial \mathcal{L}_k / \partial x^{(i)}$ is the partial derivative of the loss of the CNN-based architecture BiT-M-R101x3, and $\partial L_v / \partial x^{(i)}$ is the partial derivative of the loss of the Transformer-based architecture ViT-L/16.
Their prominence is controlled by the attacker through two manually set weighting factors, $\alpha_k$ and $\alpha_v=1-\alpha_k$.
For the ViT gradient, an additional factor is involved: the self-attention map term $\phi_v$, defined by a sum-product:
\begin{equation}
    \phi_{v}=\left(\prod_{l=1}^{n_{l}}\left[\sum_{i=1}^{n_{h}}\left(0.5 W_{l, i}^{(a t t)}+0.5 I\right)\right]\right) \odot x^{(i)}
\end{equation}
where $n_h$ is the number of attention heads per encoder block of the ViT model, $n_l$ the number of encoder blocks in the ViT model. In the ViT-L/16 for instance, $(n_h, n_l)$~=~$(16,24)$. $W_{l, i}^{(a t t)}$ is the attention weight matrix in each attention head and $I$ the identity matrix.
$\odot$ is the element wise product.

\sq{End of description !! THIS IS JUST A BLUNT COPY PASTE FOR NOW. Mention that all these attacks need backpropagating quantities somehow so this is where Pelta comes in play. Include figure here to explain the big picture of the attacks}
\begin{table}[!t]

	\normalsize
	\setlength{\tabcolsep}{7.15pt}
	\begin{center}
		\rowcolors{1}{gray!0}{gray!10}
		\begin{tabularx}{\columnwidth}{ll}
			\rowcolor{gray!25}
			\textbf{Attack} & \textbf{Parameters (CIFAR-10 and CIFAR-100)}\\
			\rowcolor{gray!1}
			FGSM  & $\epsilon$~=~$0.031$ \\
			PGD   & $\epsilon$~=~$0.031$, $\epsilon_{\text {step}}$~=~$0.00155$, steps~=~20\\
			MIM   & $\epsilon$~=~$0.031$, $\epsilon_{\text {step}}$~=~$0.00155$, $\mu$~=~1.0\\ 
               APGD  & $\epsilon$~=~0.031, $N_{\text{restarts}}$~=~1, $\rho$~=~0.75, $n^2_{\text{queries}}$~=~5$\mathrm{e}{3}$\\ 
            C\&W  & confidence~=~50, $\epsilon_{\text {step}}$~=~$0.00155$, steps~=~30\\
                SAGA & $\alpha_2$~=~$2.0\mathrm{e}{-4}$ and 0.0015, $\epsilon_{\text {step}}$~=~$3.1\mathrm{e}{-3}$ \\ 

			\rowcolor{gray!25}
			\textbf{Attack} & \textbf{Parameters (ImageNet)}\\
			\rowcolor{gray!1}
			FGSM  & $\epsilon$~=~$0.062$ \\
			PGD   & $\epsilon$~=~$0.062$, $\epsilon_{\text {step}}$~=~$0.0031$, steps~=~20\\
			MIM   & $\epsilon$~=~$0.062$, $\epsilon_{\text {step}}$~=~$0.0031$, $\mu$~=~1.0\\
               APGD   & $\epsilon$~=~$0.062$, $N_{\text{restarts}}$~=~1, $\rho$~=~0.75, $n^2_{\text{queries}}$~=~$5\mathrm{e}{3}$\\ 
                C\&W  & confidence~=~50, $\epsilon_{\text {step}}$~=~$0.0031$, steps~=~30\\
                SAGA & $\alpha_k$~=~$0.001$ and 0.0015, $\epsilon_{\text {step}}$~=~$0.0031$ \\ 
                
		\end{tabularx}
  
  \rule{250pt}{0.7pt}
	\end{center}
	\caption{\label{tab:params}Attack parameters.}
	\label{table:params}
\end{table}

\begin{center}
\begin{table*}[!t]
	\centering
	\normalsize
	\setlength{\tabcolsep}{6pt}
	\begin{center}
		\rowcolors{1}{gray!0}{gray!10}
		\begin{tabularx}{1.89\columnwidth}{lccccccccccc}
			\rowcolor{gray!1}
			\rowcolor{gray!50}
			\textbf{CIFAR-10} & \multicolumn{2}{c|}{{\textbf{FGSM}}} & \multicolumn{2}{c|}{{\textbf{PGD}}} & \multicolumn{2}{c|}{{\textbf{MIM}}} & \multicolumn{2}{c|}{{\textbf{C\&W}}} & \multicolumn{2}{c|}{{\textbf{APGD}}} & \multicolumn{1}{c}{{\textbf{Clean}}}\\
\rowcolor{gray!1}
ViT-L/16 &   57.2\% & \multicolumn{1}{c|}{99.3\%} & 1.2\% & \multicolumn{1}{c|}{99.2\%} & 6.2\% & \multicolumn{1}{c|}{98.7\%} & 0.0\% & \multicolumn{1}{c|}{99.1\%} & 0.0\%& \multicolumn{1}{c|}{90.6\%} & 99.4\%\\
ViT-B/16 &   38.6\% & \multicolumn{1}{c|}{91.2\%} & 0.0\% & \multicolumn{1}{c|}{96.2\%} & 0.3\% & \multicolumn{1}{c|}{97.3\%} & 0.0\% & \multicolumn{1}{c|}{95.8\%} & 0.0\%& \multicolumn{1}{c|}{88.9\%} & 98.5\%\\
ViT-B/32 &   33.5\% & \multicolumn{1}{c|}{92.8\%} & 2.1\% & \multicolumn{1}{c|}{98.0\%} & 3.3\% & \multicolumn{1}{c|}{97.9\%} & 0.0\% & \multicolumn{1}{c|}{96.9\%} & 0.0\%& \multicolumn{1}{c|}{89.9\%} & 98.0\%\\
ResNet-56&   23.9\% & \multicolumn{1}{c|}{75.6\%} & 0.0\% & \multicolumn{1}{c|}{81.3\%} & 0.0\% & \multicolumn{1}{c|}{95.9\%} & 0.0\% & \multicolumn{1}{c|}{95.2\%} & 0.0\%& \multicolumn{1}{c|}{57.1\%} & 93.0\%\\
ResNet-164&  30.0\% & \multicolumn{1}{c|}{78.7\%} & 0.0\% & \multicolumn{1}{c|}{82.7\%} & 0.0\% & \multicolumn{1}{c|}{96.3\%} & 0.0\% & \multicolumn{1}{c|}{96.0\%} & 0.0\%& \multicolumn{1}{c|}{60.0\%} & 93.9\%\\
BiT-M-R101x3&84.8\% & \multicolumn{1}{c|}{90.9\%} & 0.0\% & \multicolumn{1}{c|}{74.1\%} & 0.0\% & \multicolumn{1}{c|}{83.1\%} & 0.0\% & \multicolumn{1}{c|}{98.0\%} & 0.0\%& \multicolumn{1}{c|}{57.3\%} & 98.9\%\\			

\rowcolor{gray!50}
\textbf{CIFAR-100} & \multicolumn{2}{c|}{{\textbf{FGSM}}} & \multicolumn{2}{c|}{{\textbf{PGD}}} & \multicolumn{2}{c|}{{\textbf{MIM}}} & \multicolumn{2}{c|}{{\textbf{C\&W}}} & \multicolumn{2}{c|}{{\textbf{APGD}}} & \multicolumn{1}{c}{{\textbf{Clean}}}\\
\rowcolor{gray!1}
ViT-L/16 &   30.1\% & \multicolumn{1}{c|}{99.2\%} & 1.0\% & \multicolumn{1}{c|}{98.9\%} & 4.0\% & \multicolumn{1}{c|}{98.0\%} & 0.0\% & \multicolumn{1}{c|}{99.0\%} & 0.0\%& \multicolumn{1}{c|}{90.0\%} & 93.9\%\\
ViT-B/16 &   19.3\% & \multicolumn{1}{c|}{91.1\%} & 2.7\% & \multicolumn{1}{c|}{93.2\%} & 0.9\% & \multicolumn{1}{c|}{96.9\%} & 0.0\% & \multicolumn{1}{c|}{97.9\%} & 0.0\%& \multicolumn{1}{c|}{88.3\%} & 92.5\%\\
ViT-B/32 &   20.0\% & \multicolumn{1}{c|}{92.9\%} & 2.9\% & \multicolumn{1}{c|}{92.0\%} & 1.9\% & \multicolumn{1}{c|}{98.4\%} & 0.0\% & \multicolumn{1}{c|}{96.2\%} & 0.0\%& \multicolumn{1}{c|}{89.0\%} & 93.5\%\\
ResNet-56&   5.2\%  & \multicolumn{1}{c|}{81.5\%} & 0.1\% & \multicolumn{1}{c|}{82.6\%} & 3.3\% & \multicolumn{1}{c|}{95.1\%} & 0.0\% & \multicolumn{1}{c|}{96.0\%} & 0.0\%& \multicolumn{1}{c|}{60.8\%} & 70.0\%\\
ResNet-164&  7.9\%  & \multicolumn{1}{c|}{83.8\%} & 0.2\% & \multicolumn{1}{c|}{83.7\%} & 3.9\% & \multicolumn{1}{c|}{97.6\%} & 0.0\% & \multicolumn{1}{c|}{94.2\%} & 0.0\%& \multicolumn{1}{c|}{62.1\%} & 76.1\%\\
BiT-M-R101x3&3.3\%  & \multicolumn{1}{c|}{85.4\%} & 0.0\% & \multicolumn{1}{c|}{75.7\%} & 0.0\% & \multicolumn{1}{c|}{82.9\%} & 0.0\% & \multicolumn{1}{c|}{98.8\%} & 0.0\%& \multicolumn{1}{c|}{59.8\%} & 90.2\%\\
\rowcolor{gray!50} \textbf{ImageNet} & \multicolumn{2}{c|}{{\textbf{FGSM}}} & \multicolumn{2}{c|}{{\textbf{PGD}}} & \multicolumn{2}{c|}{{\textbf{MIM}}} & \multicolumn{2}{c|}{{\textbf{C\&W}}} & \multicolumn{2}{c|}{{\textbf{APGD}}} & \multicolumn{1}{c}{{\textbf{Clean}}}\\
\rowcolor{gray!1} 
ViT-L/16 &     27.5\% & \multicolumn{1}{c|}{94.0\%} & 0.0\% & \multicolumn{1}{c|}{93.9\%} & 0.0\% & \multicolumn{1}{c|}{97.4\%}    & 0.0\%   & \multicolumn{1}{c|}{99.3\%}   & 0.0\%  & \multicolumn{1}{c|}{86.5\%}   & 82.7\%\\
ViT-B/16 &     22.1\% & \multicolumn{1}{c|}{92.1\%} & 0.0\% & \multicolumn{1}{c|}{92.0\%} & 0.0\% & \multicolumn{1}{c|}{97.4\%} & 0.0\% & \multicolumn{1}{c|}{99.1\%} & 0.0\%& \multicolumn{1}{c|}{87.1\%} & 80.0\%\\
BiT-M-R101x3 & 24.2\% & \multicolumn{1}{c|}{83.8\%} & 0.0\% & \multicolumn{1}{c|}{76.8\%} & 0.0\% & \multicolumn{1}{c|}{83.2\%} & 0.0\% & \multicolumn{1}{c|}{98.2\%} & 0.0\%& \multicolumn{1}{c|}{73.4\%} & 79.3\%\\
BiT-M-R152x4 & 67.0\% & \multicolumn{1}{c|}{85.8\%} & 0.0\% & \multicolumn{1}{c|}{87.1\%} & 0.0\% & \multicolumn{1}{c|}{93.7\%} & 0.0\% & \multicolumn{1}{c|}{98.0\%} & 0.0\%& \multicolumn{1}{c|}{67.1\%} & 85.1\%\\

			\specialrule{.8pt}{0pt}{2pt}
		\end{tabularx}
	\end{center}
	\caption{\label{tab:PGDresults}Robust accuracy of non-shielded (left) versus shielded (right) individual models against a benchmark of five white-box attacks on CIFAR-10, CIFAR-100 and ImageNet (higher values favor the defender). Clean accuracy over 1000 random validation samples is provided for illustrative purpose.}
	\label{table:PGDresults}
\end{table*}
\end{center}

\begin{table*}[!t]
	\centering
	\normalsize
	\setlength{\tabcolsep}{10pt}
	\begin{center}
		\rowcolors{1}{gray!0}{gray!10}
		\begin{tabularx}{1.9\columnwidth}{Xcccccc}
			\rowcolor{gray!50}
			\textbf{CIFAR-10} & \multicolumn{2}{c|}{\textbf{Baseline}} & \multicolumn{4}{c}{\textbf{Applied Shield}}\\
			\rowcolor{gray!25}
			\textbf{Model Acc.} & \textbf{Clean} & \multicolumn{1}{c|}{\textbf{Random}} & \textbf{None} & \textbf{ViT-L/16} & \textbf{BiT-M-R101x3} & \textbf{Ensemble}\\
			\rowcolor{gray!2}
			ViT-L/16  &   99.4\% & \multicolumn{1}{c|}{99.5\%} & 28.1\% & 99.2\% & 14.1\% & 99.5\%\\
			BiT-M-R101x3   & 98.8\% & \multicolumn{1}{c|}{98.8\%} & 25.2\% & 0.3\% & 78.9\% & 98.5\%\\
			Ensemble  &   99.1\% & \multicolumn{1}{c|}{98.9\%} & 27.2\% & 49.7\% & 46.4\% & 98.8\%\\
			
			\rowcolor{gray!50} \textbf{CIFAR-100} & \multicolumn{2}{c|}{\textbf{Baseline}} & \multicolumn{4}{c}{\textbf{Applied Shield}}\\
			\rowcolor{gray!25}
			\textbf{Model Acc.} & \textbf{Clean} & \multicolumn{1}{c|}{\textbf{Random}} & \textbf{None} & \textbf{ViT-L/16} & \textbf{BiT-M-R101x3} & \textbf{Ensemble}\\
			\rowcolor{gray!2}
			ViT-L/16  &    94.0\% & \multicolumn{1}{c|}{99.4\%} & 5.2\%  & 99.6\% & 4.2\%  & 99.8\%\\
			BiT-M-R101x3 & 89.9\% & \multicolumn{1}{c|}{98.3\%} & 18.3\% & 0.2\%  & 50.0\% & 82.2\%\\
			Ensemble  &    92.0\% & \multicolumn{1}{c|}{98.9\%} & 12.2\% & 49.5\% & 27.5\% & 90.8\%\\
			
			\rowcolor{gray!50} \textbf{ImageNet} & \multicolumn{2}{c|}{\textbf{Baseline}} & \multicolumn{4}{c}{\textbf{Applied Shield}}\\
			\rowcolor{gray!25}
			\textbf{Model Acc.} & \textbf{Clean} & \multicolumn{1}{c|}{\textbf{Random}} & \textbf{None} & \textbf{ViT-L/16} & \textbf{BiT-M-R101x3} & \textbf{Ensemble}\\
			\rowcolor{gray!2}
			ViT-L/16  &    82.6\% & \multicolumn{1}{c|}{100.0\%} & 6.3\%  & 99.2\% & 6.1\%  & 97.5\%\\
			BiT-M-R152x4 & 85.6\% & \multicolumn{1}{c|}{100.0\%} & 15.2\% & 0.6\%  & 45.5\% & 76.2\%\\
			Ensemble  &    84.3\% & \multicolumn{1}{c|}{100.0\%} & 10.8\% & 49.9\% & 25.8\% & 87.0\%\\
			\specialrule{.8pt}{0pt}{2pt}
		\end{tabularx}
	\end{center}
	\caption{\label{tab:SAGAresults}Robust accuracy of a shielded ensemble against SAGA on 1000 corretly classified CIFAR-10, CIFAR-100 and ImageNet samples (higher values favor the defender). Baseline values show clean accuracy, astuteness against random uniform attack on the $l_{\infty}$ ball. Applied Shield values show per-model robust accuracy against different shielding setups.}
	\label{table:SAGAresults}
\end{table*}

Facing the \sys shielded setting, the attacker carries out the SAGA without the masked set $\left\{\partial f/\partial {x}\right\}^L$ (and the adjacent quantities otherwise enabling its unambiguous reconstruction).
It tries to exploit the adjoint $\delta_{L+1}$ of the last clear layer by applying to it a random-uniform initialized upsampling kernel.
This process, called \textit{transposed convolution}~\cite{Dumoulin2016AGT}, essentially is a geometrical transformation applied to the vector gradient at the backward pass of a convolutional layer.
Although this method does not offer any guarantee of convergence towards a successful adversarial example, it allows to understand whether the adjoint can still serve as a last resort when no priors on the shielded parts are available under limited resource constraint. Individual models are attacked in a similar manner, replacing gradient terms of the shielded computations of the defender by gradients of a substitute transposed convolution.
We will therefore ask: in the absence of the shielded quantities, \emph{does upsampling constitute a last resort for the attacker?}

\subsection{Benchmarks and results}
We select 1000 correctly classified random samples from  CIFAR-10, CIFAR-100 and the ImageNet Large Scale Visual Recognition Challenge (ILSVRC)~\cite{Russakovsky2015Dec}, a subset of ImageNet-21K. This means that the robust accuracy over these samples is 100\% if no attack is run.
We evaluate the average robust accuracy of individual models against five attacks in a setting where the model is not shielded and a setting where the model is shielded.
The ensemble model is evaluated against the SAGA in four settings: no model is shielded, only the BiT model is shielded, only the ViT model is shielded, both models are shielded (maximum protection for the ensemble). 
For reference, the clean accuracy of the models over 1000 random samples from the validation set of each dataset is provided. Attack parameters are provided in Table~\ref{tab:params}.

\begin{figure}[!t]
	\begin{center}
		\includegraphics[scale=0.36,trim={0 0 0 0}]{{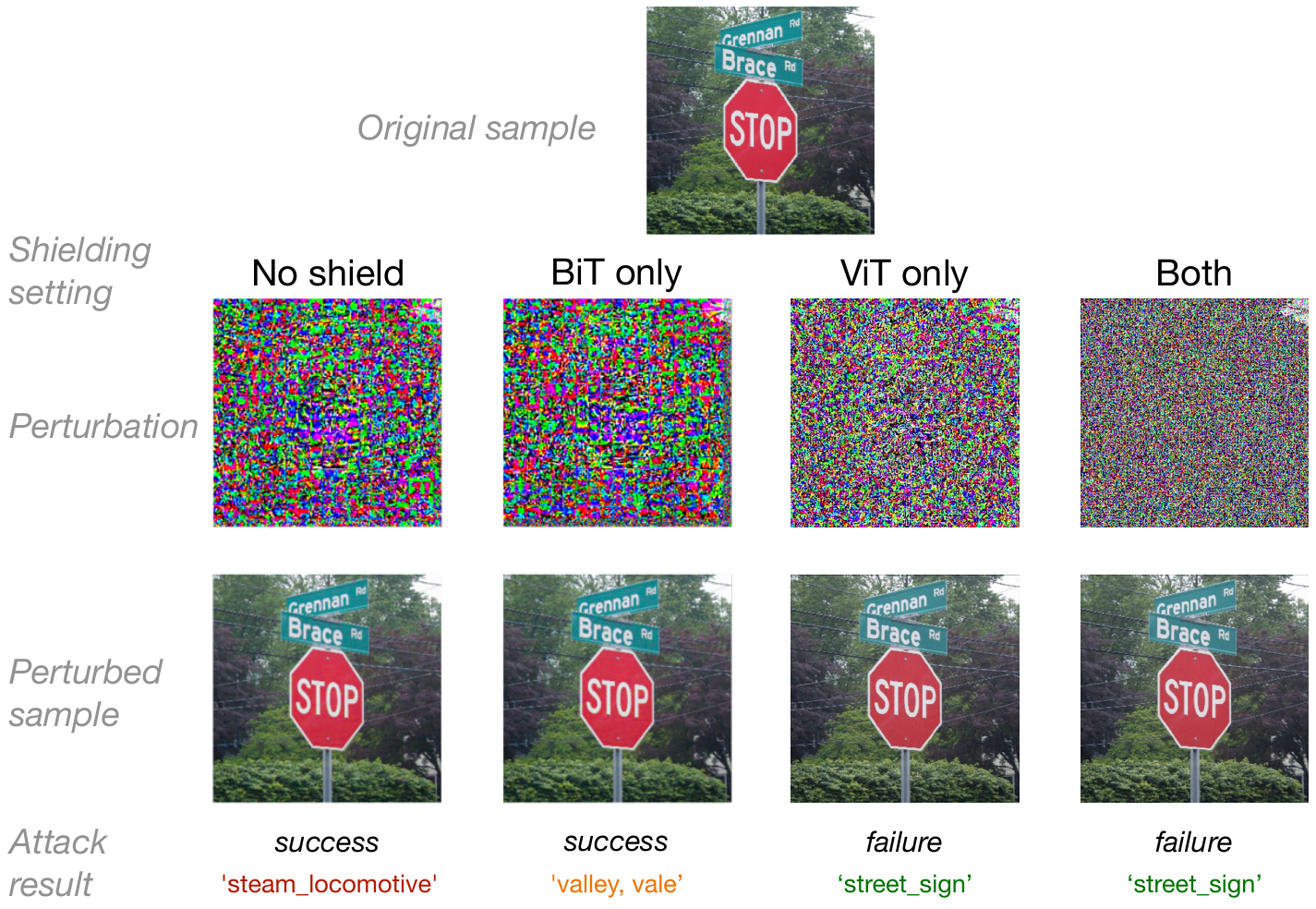}}
	\end{center}
	\caption{SAGA adversarial samples in four different shielding settings from a correctly classifed sample. \label{fig:advex}}
\end{figure}

Table~\ref{table:PGDresults} shows our results for the individual models against the five attacks and Table~\ref{table:SAGAresults} shows our results for the ensemble model against the SAGA.
For illustrative purposes, Fig.\ref{fig:advex} shows the generated perturbation on one sample in the four settings of the ensemble model against the SAGA.

\textit{Does \sys mitigate white-box attacks?}
We observe that the shielding greatly preserves the astuteness of individual models and of the ensemble, with up to $98.8\%$ robust accuracy for the ensemble ($1.2\%$ attack success rate) comparable to random uniform pixel modifications, and up to $99.3\%$ robust accuracy for individual models.
It can be noted that, generally, the size of the model has a positive influence on the astuteness after shielding across attacks; however, this effect seems largely overwhelmed by the advantage variants of ViT have over CNN-based models before and after shielding.
Additionally, we see that for the ensemble defense, individual model robust accuracies are not equally protected: the ViT model benefits more from \sys when applied only to it than BiT.
We explain these results as a general advantage in robustness of Transformer-based architectures over CNNs~\cite{DBLP:journals/corr/abs-2110-02797} with BiT being more sensitive to targeted attacks than its counterpart, and also more sensitive to adversarial examples crafted against the ViT. 
We further note that in the ensemble, individual robust accuracies worsen compared to the non-shielded setting when only their counterpart is shielded. 
The reason for this is attributed to the fact that SAGA solely directs the sample towards augmenting the non-shielded loss, while disregarding the shielded loss, consequently resulting in the creation of adversarial samples that exclusively aim to exploit vulnerabilities in the non-shielded model.

\textit{Does upsampling constitute a last resort for the attacker?} Interestingly, for the ensemble, the attack success rate of the upsampling against the \sys defense scheme sometimes surpasses that of the random uniform attack against BiT when the shield is applied only to it.
We explain this behaviour as follows: contrary to the shielded layer in ViT that projects the input onto an embedding space, the last clear layer adjoint $\delta_{L+1}$ in the BiT still carries spatial information that could be recovered \eg through average upsampling.
These results suggest shielding both models for optimal security, given sufficient encrypted memory available in a target TEE.

A similar remark can be made about the behavior of CNN-based models in the individual attacks.
While ViT variants generally fare very well against most attacks to the exception of the most sophisticated (APGD), ResNets and BiTs have overall low astuteness, indicating the attacks behave largely better than random. This suggests that larger parts of the model should be included in the enclave of the \sys scheme to mitigate the effectiveness of the upsampling by the attacker.

\break
\section{System Implications}
\label{section:discuss}
   As previously mentioned, TEEs typically operate in a secure mode that is designed to provide protection against external attacks. This secure mode can add overhead and complexity to communication between the secure world and the rest of the system, which can in turn impact data throughput. For example, when data is transferred between the TEE and the non-secure world, it may need to be encrypted and decrypted, which can slow down communication. Data transferred between the secure world and the TEE traditionally requires secure communication protocols to prevent unauthorized access or tampering, thus hindering the velocity of the data transfer process, as encryption and decryption may be necessary to protect the data. Moreover, a context switch is typically required to move from one execution context to another. This context switching can introduce additional overhead and slow down the data transfer process.

  Because \sys is designed to hide sensitive data at inference time through the use of a TEE, a throughput bottleneck could be felt at two stages of the federated endeavour. The first case is the most self-explanatory: even after deployment following FL rounds, inference with \sys still supposes parts of the model are inside a TEE and sensitive operations are carried inside the enclave. This requires context switches and establishing a secure communication channel between worlds to either feed the first computation nodes of the model with the input data or extracting the output of the last shielded layer to carry on subsequent operations with the clear and deeper segment of the model. Fortunately, these elementary TEE methods usually range from microseconds up to milliseconds at most for either SGX or TrustZone~\cite{Mukherjee2019Nov, Weisse2017Jun}, which is commensurate with the real-time usage of most current edge ML applications that fit into FL (text prediction, sentiment analysis, health monitoring, speech recognition etc.) as recently demonstrated~\cite{Babar2022Nov}. As a rule, it should be kept in mind that minimizing context switches is an important optimization technique in the design of TEE applications.

   The second case is the training phase of the FL scheme. During this phase, the use of an optimization algorithm puts more strain on the TEE and its communication channels. For instance, inside the enclave, gradients which were not generated during regular end-user inference are now being computed: these gradients seldom need to be read from within the enclave in order to be sent for aggregation, which adds in bandwidth overhead. However, many of these limitations are taken into account when tuning the parameters of the protocol for each FL round. For example, the frequency at which the weight updates are pulled out of the enclave to be sent to the aggregating server could be lowered to allow averaging hidden gradients over larger batches on the client nodes. Overall, training protocols of an FL scheme are expected to harness the idle state of edge devices to handle intermittent compute node availability~\cite{Yan2020DistributedNO}. The extra bandwidth overhead of the second case should therefore not impact user experience as much as it does the strategy of the FL training rounds.

\section{Conclusion and Future Work}
\label{section:conclu}
In federated learning, adversarial attacks are the basis of several trojaning and poisoning attacks. However, they are difficult to defend against at inference time under the white-box hypothesis, which is the default setting in FL. 
We described how to mitigate such attacks by using hardware obfuscation to sidestep the white-box. We introduce a novel defense, \sys: a lightweight shielding method which relies on TrustZone enclaves to mask few carefully chosen parameters and gradients against iterative gradient-based attacks. 
Our evaluation shows promising results, sensibly mitigating the effectiveness of state-of-the-art attacks. To the best of our knowledge, \sys also constitute the first attempt at defending against the Self-Attention Gradient Attack. 

We intend to extend this work along the following directions.
Because the use of enclaves calls for somewhat costly normal-world to secure-world communication mechanisms, properly evaluating the speed of each collaborating device under our shielding scheme is needed on top of considering the aforementioned memory overheads (Table~\ref{table:size}) in order to assess the influence of \sys on the FL training's practical performance.
Additionally, a natural extension to this work is to apply \sys along with existing software defenses~\cite{REN2020346} to assess their combined benefits against a sophisticated attacker. 

As mentioned in \S\ref{subsection:bpda}, it should also be explored that an attacker can \emph{(i)} exploit commonly used embedding matrices and subsequent parameters across existing models as a prior on the shielded layers (this case being circumvented by the defender if it trains its own first parameters) or \emph{(ii)} to train on their own premises  the aforementioned $g$ backward approximation (which needs not be of the same architecture as the shielded layers), although recent work shows limitation for such practice~\cite{sitawarin2022demystifying}.
\balance

\bibliographystyle{IEEEtran}
\bibliography{paper}

\end{document}